\def\year{2022}\relax
\documentclass[letterpaper]{article} 
\usepackage{aaai22}  
\usepackage{times}  
\usepackage{helvet}  
\usepackage{courier}  
\usepackage[hyphens]{url}  
\usepackage{graphicx} 
\usepackage{amsmath}
\usepackage{natbib}  
\usepackage{caption} 
\DeclareCaptionStyle{ruled}{labelfont=normalfont,labelsep=colon,strut=off} 
\usepackage{multirow}
\usepackage{makecell}
\usepackage{booktabs} 
\usepackage{algorithm}
\usepackage{algorithmic}
\usepackage{newfloat}
\usepackage{listings}
\floatstyle{ruled}
\newfloat{listing}{tb}{lst}{}
\floatname{listing}{Listing}
\title{SVT-Net: Super Light-Weight Sparse Voxel Transformer for Large \\Scale Place Recognition}
\author{ Zhaoxin Fan\textsuperscript{\rm 1}, Zhenbo Song\textsuperscript{\rm 3}, Hongyan Liu \textsuperscript{\rm 4}\footnote{Corresponding authors}, Zhiwu Lu\textsuperscript{\rm 2}, Jun He\textsuperscript{\rm 1}\footnotemark[1] , and Xiaoyong Du\textsuperscript{\rm 1}
}
\affiliations{
    \textsuperscript{\rm 1}  Key Laboratory of Data Engineering and Knowledge Engineering of MOE\\
School of Information, Renmin University of China, 100872, Beijing, China\\
    
        \textsuperscript{\rm 2}  Gaoling School of Artificial Intelligence, Renmin University of China, 100872, Beijing, China\\
      
        \textsuperscript{\rm 3}  School of Computer Science and Engineering, Nanjing University of Science and Technology, 210094, Nanjing, China\\
                \textsuperscript{\rm 4}  Department of Management Science and Engineering, Tsinghua University, 100084, Beijing, China\\
                \{fanzhaoxin, luzhiwu, hejun\}@ruc.edu.cn, hyliu@tsinghua.edu.cn, songzb@njust.edu.cn \\
}

\usepackage{bibentry}

\begin{document}

\maketitle

\begin{abstract}
   Simultaneous Localization and Mapping (SLAM) and Autonomous Driving are becoming increasingly more important in recent years. Point cloud-based large scale place recognition is the spine of them. While many models have been proposed and have achieved acceptable performance by learning short-range local features,  they always skip long-range contextual properties. Moreover, the model size also becomes a serious shackle for their wide applications. To overcome these challenges, we propose a super light-weight network model termed SVT-Net. On top of the highly efficient 3D Sparse Convolution (SP-Conv), an Atom-based Sparse Voxel Transformer (ASVT) and a Cluster-based Sparse Voxel Transformer (CSVT) are proposed respectively to learn both short-range local features and long-range contextual features. Consisting of ASVT and CSVT, SVT-Net can achieve state-of-the-art performance in terms of both recognition accuracy and running speed with a super-light model size (0.9M parameters). Meanwhile, for the purpose of further boosting efficiency, we introduce two simplified versions, which also achieve state-of-the-art performance and further reduce the model size to 0.8M and 0.4M respectively.
\end{abstract}

\section{Introduction}

\begin{figure}[ht]
\centering  
\includegraphics[width=0.4\textwidth]{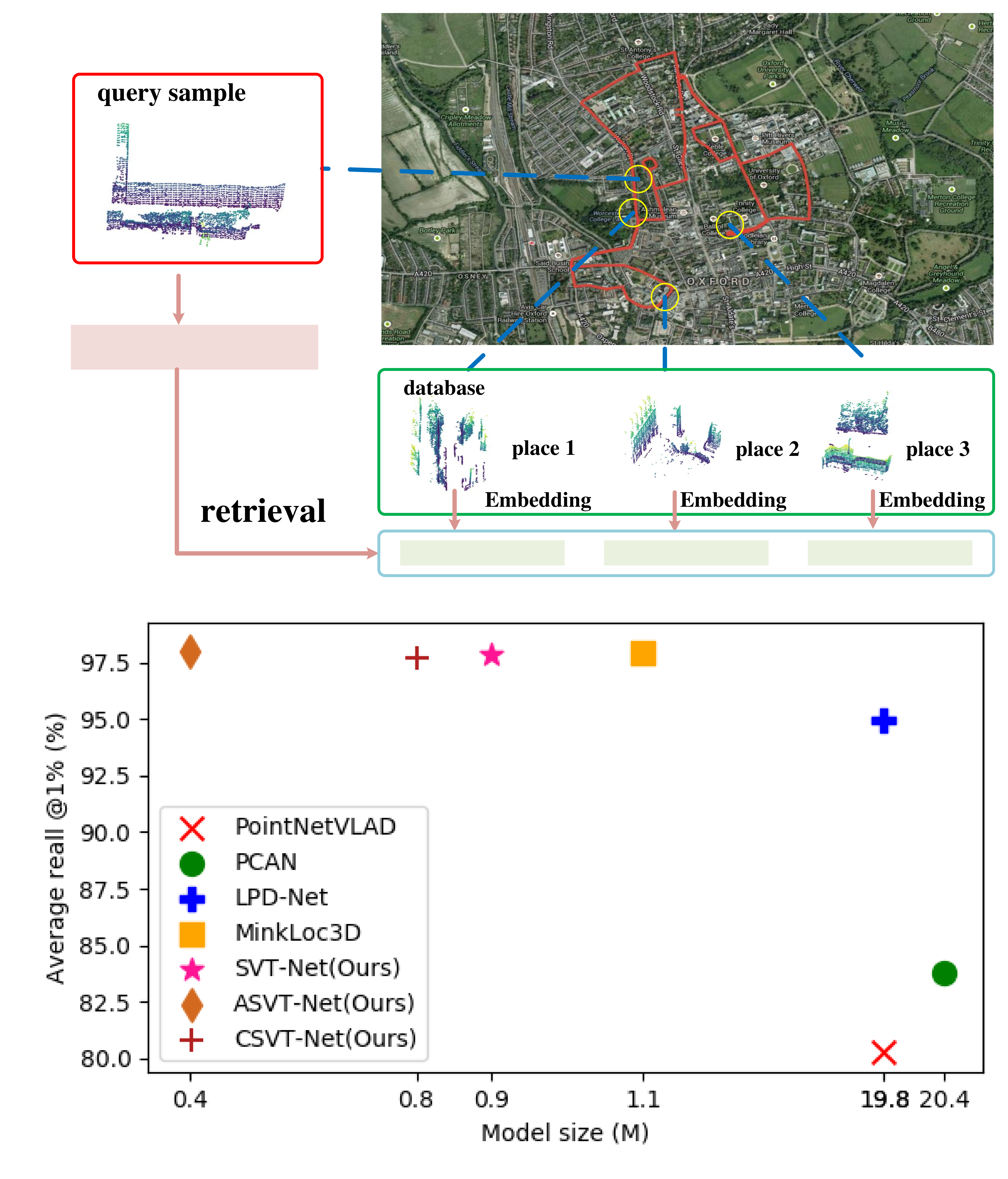}  

\caption{(Top) Pipeline of point cloud based place recognition. (Bottom)  Model size and accuracy.}

\label{task_and_size}
\end{figure}

Large scale place recognition is the spine of a wide range of applications like Simultaneous Localization and Mapping (SLAM) \cite{mur2017orb}, Autonomous Driving \cite{levinson2011towards}, Robot Navigation \cite{ravankar2018path}, etc. Commonly, the place recognition result can be used for loop-closure \cite{chen2020overlapnet} in a SLAM system or for user location in a indoor vision positioning system, when GPS signal is not available. Fig. \ref{task_and_size} (Top) illustrates a common pipeline of large place recognition. For a large scale region, a database of scenes (usually represented by point clouds or images )  tagged with UTM coordinates acquired from GPS/INS readings are constructed in advance. When a user traverses the same region, he/her may collect a query scene from scratch. Then the most similar scene to the query scene should be retrieved from the database to determine where the location of the query scene is.

A straight-forward idea for this task is to use images to learn global descriptors for  accurate and efficient scene retrieval \cite{li2010location,han2017sral,yu2019spatial}.  However, images are sensitive to illumination, weather change, diurnal variation, etc, making  models based on them unstable  and unreliable.  Besides, images are short of perceiving 3D scenes due to lack of depth information. Recently, a line of  point cloud based deep learning models \cite{uy2018pointnetvlad,zhang2019pcan,sun2020dagc,liu2019lpd,fan2020srnet,xia2021soe,komorowski2021minkloc3d} for large scale place recognition have been proposed.  Since point clouds are invariant to illumination and weather changes,  point cloud based methods are more robust than image based methods. Besides, since point clouds contain richer 3D information, global descriptors learned from them are stronger in describing 3D scenes than image descriptors and therefore they always achieve better performance. 

 Though better, existing point cloud based methods still face three main challenges. 1) Most of existing methods learn descriptors from point-wise point cloud encoders, which are sensitive to local noise. These local noise may stand for scene details and being important for some fine grained level tasks such as segmentation. However, they are useless for place recognition but even become a burden for the network to understand the scene.  Therefore, they should be regarded as noise and outliers. 2) We observe that most of previous methods only consider how to better extract short-range local features, while the equally important long-range contextual properties have long been skipped. And we argue that lacking awareness of long-range contextual properties, power of the learned descriptors would be greatly limited. 3) Most of  existing models are suffered from huge model size, which stops their application in resource constrained portable devices. Considering the above issues, we claim that designing a  local noise-insensitive light-weight point cloud descriptor extraction model that can capture long-range contextual features is necessary.

In this paper, we propose a novel super light-weight network named SVT-Net for point cloud based large scale place recognition. SVT-Net's nework architecture is built upon the delicate light weight 3D Sparse Convolution (SP-Conv) \cite{choy20194d}. The reason why we choose SP-Conv lies in two aspects. First, the sparse voxel representation require to voxelize point cloud, which reduces local noise but retains most of overall scene geometries. Therefore, it can liberate the model from understanding useless scene details. Second, the SP-Conv is efficient and fast. It only computes outputs for predefined coordinates and saves them into a compact sparse tensor. In other words, it meets our requirements for building a light-weight model.  

However, simply stacking SP-Conv layers may cause neglect of long-range contextual properties. A direct way to solve this problem is introducing Vision Transformers  \cite{dosovitskiy2020image} for learning long-range contextual features. There indeed exists point cloud Transformers \cite{guo2020pct} in literature, however, they are not suitable for the point cloud based place recognition task. It is because all existing point cloud Transformers are point-wise modules and therefore not efficient enough. Besides, as mentioned before, point-wise modules may suffer from local noise. Therefore, we propose two kinds of Sparse Voxel Transformers (SVTs) tailored for large scale place recognition on top of SP-Conv layers named Atom-based Sparse Voxel Transformer (ASVT) and Cluster-based Sparse Voxel Transformer (CSVT) respectively.  ASVT and CSVT implicitly extract long-range contextual features  from the sparse voxel representation through two perspectives: attending on different key atoms and clustering different key regions in the feature space, thereby helping to obtain more discriminative descriptors through interacting different atoms (to learn inter-atoms long-range features) and different clusters (to learn inter-clusters long-range features) respectively. Since SP-Conv only conducts convolution operation on non-empty voxels, it is computational efficient and flexible, so do the two SVTs built upon it. Thanks to the strong capabilities of the two SVTs,  our proposed model can learn sufficiently powerful descriptors from an extremely shallow network architecture. And thanks to the shallow network architecture, model size of SVT-Net is very small as shown in Fig. \ref{task_and_size} (Bottom).

We conduct extensive experiments on Oxford RobotCar dataset \cite{maddern20171} and three in-house datasets \cite{uy2018pointnetvlad} to verify the effectiveness and efficiency of SVT-Net. Results show that though light-weight, SVT-Net can achieve state-of-the-art performance in terms of both accuracy and speed. What's more, to further increase speed and reduce model size, we introduce two simplified version of SVT-Net: ASVT-Net and  CSVT-Net, which also achieve state-of-the-art performances with further reduced model sizes of only 0.8M parameters and 0.4M parameters respectively. 

Our main contributions are three folds. 1) We propose a novel light-weight point cloud based place recognition model named SVT-Net as well as two simplified versions:  ASVT-Net and  CSVT-Net, which all achieve state-of-the-art performance in terms of both accuracy and speed with a extremely small model size. 2) We propose Atom-based Sparse
Voxel Transformer (ASVT) and Cluster-based Sparse Voxel Transformer (CSVT) for learning long-range contextual features hidden in point clouds. To the best of our knowledge, we are the first to propose Transformers for sparse voxel representations. 3) We have conducted extensive quantitative and qualitative experiments to verify the effectiveness and efficiency of our proposed models and analysed what the two proposed Transformers actually learn. 
\section{Related Work}

\subsection{Large Scale Place Recognition}

Large scale place recognition plays an important role in SLAM and autonomous driving and has been interested in by many researchers for a  long time. In early years, hand-craft features \cite{galvez2012bags,fernandez2013fast,johns2011images} or learned features \cite{arandjelovic2016netvlad,yu2019spatial,hausler2021patch} extracted from images are used for place recognition.  These methods, though straight-forward, are  suffered from vulnerability of features caused by images' sensitivity towards illumination, weather change, diurnal variation, etc.

Compared to image, point cloud is more insensitive to environmental changes, therefore it is a better alternative for place recognition. PointNetVLAD \cite{uy2018pointnetvlad} adopts PointNet \cite{qi2017pointnet} and NetVLAD \cite{arandjelovic2016netvlad} to learn global point cloud descriptors for this task.
Then, a series of following works \cite{zhang2019pcan,sun2020dagc,fan2020srnet,liu2019lpd,xia2021soe,komorowski2021minkloc3d} are proposed. They use graph networks, attentions and voxel representation to learn powerful global descriptors for this task respectively. However, most of them are suffered from three aspects: first, they fail to learn long-range contextual features of  scenes from point cloud; second, model size and efficiency are not considered in their methods; third, they are sensitive to local noise. In our work, we design two light-weight but strong Sparse Voxel Tranformers to tackle the above problems.

\subsection{Vision Transformers}

Transformer \cite{vaswani2017attention} is one of the most successful design for  natural language processing (NLP)\cite{devlin2018bert, hu2018squeeze, yang2019xlnet, kim2021vilt, chen2021crossvit,chen20182}, the core of which is a self-attention mechanism  to capture long-range contextual features.  Recently, inspired by the great success of Transformer in NLP, researchers begin to design Transformers tailored for computer vision tasks.

Therefore, Vision Transformer (ViT) \cite{dosovitskiy2020image} is proposed recently. It adopts the  idea of self-attention and divides images to 16x16 visual words. In this way, images can be processed like nature language. Then, a variety of following works \cite{wu2020visual,wang2021pyramid,liu2021swin,jiang2021transgan} are proposed based on it.  However, all the above introduced vision Transformers are designed for learning from images.  To boost the performance of point cloud based tasks,  point-wise vision  Transformers like \cite{zhao2020point,guo2020pct} are proposed. Though tailored for point clouds, they are not suitable for the place recognition task. Because they are not light-weight enough and are suffered from small local noise in raw point clouds. In contrast, we propose two kinds of super-light Sparse Voxel Transformers to learn global features from scenes, which are less suffered from local noise and are much more efficient. To our knowledge, this is the first work designs Sparse Voxel Transformers for point clouds.

\section{Methodology}

\subsection{Problem Definition}
Let $M_r={\{m_i| i = 1,2,...,M\}}$ be a database of pre-defined 3D submaps (represented as point clouds), and Q be a query point cloud scan. The place recognition problem is defined as retrieving a submap $m_s$ from $M_r$ with the goal of $m_s$ is the closest one to Q. To achieve accurate retrieving, a deep learning model $F(*)$ that can embed all point clouds into discriminative global descriptors, e.g. $Q \to f_q \in R^d $, is required so that a following KNNs algorithm can be used for finding $m_s$. 

To meet the goal, we choose to use the sparse voxel representation of point cloud as input and choose 3D Sparse Convolution (SP-Conv) \cite{choy20194d} as the basic unit to build the deep learning model. To employ SP-Conv, we first voxelize  all point clouds into sparse voxel representations, e.g.  $Q \to Q^v \in R^{L \times W \times H \times 1}$, where for each voxel, 1 means that it is occupied by any points in $Q$, called non-empty voxel, and otherwise 0, called empty voxel.  SP-Conv operation is only conducted on non-empty voxels. Hence, it is very efficient and flexible. Next, we will introduce the two proposed Transformers: the Atom-based Sparse Voxel Transformer (ASVT) and the Cluster-based Sparse Voxel Transformer (CSVT) respectively. And then, the overall network architecture of SVT-Net as well as network architectures of the two simplified versions (ASVT-Net and CSVT-Net) will be introduced in detail. The loss function will be presented finally.

\begin{figure}[t]
\centering  
\includegraphics[width=0.48\textwidth]{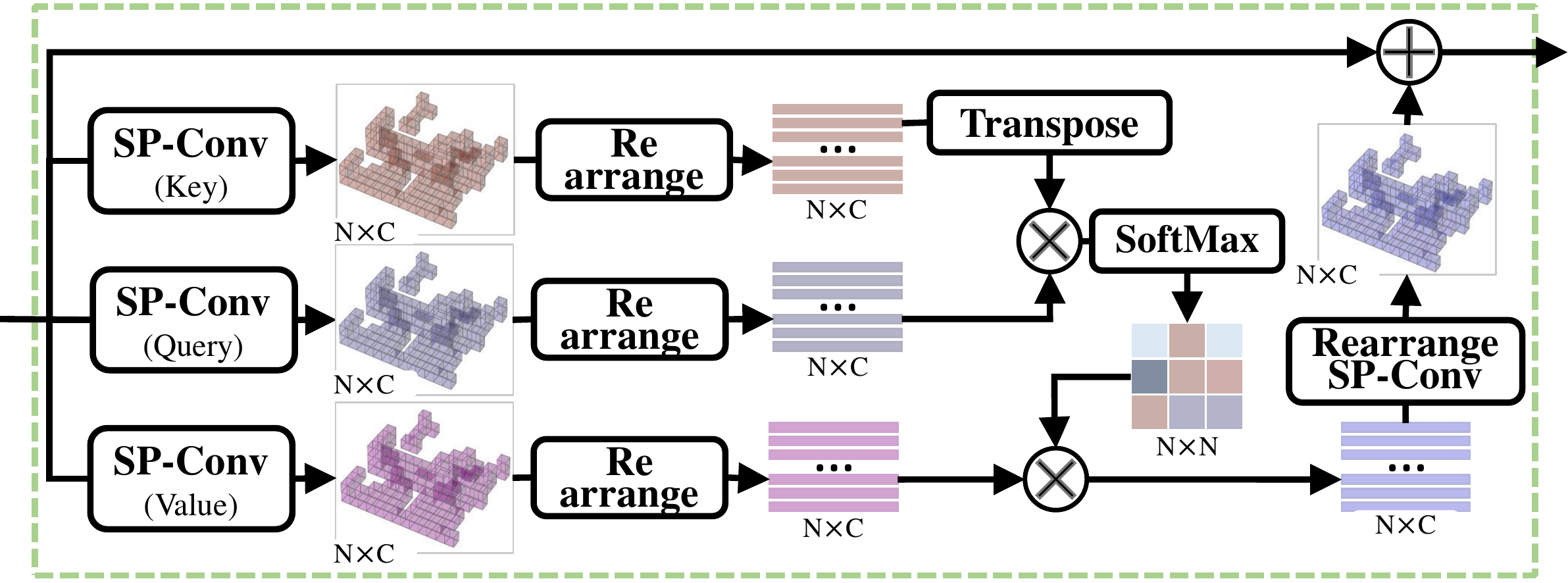}  
\caption{Network architecture of ASVT.} 
\label{ASVT}
\end{figure}

\subsection{Atom-based Sparse Voxel Transformer}

As mentioned before, simply stacking SP-Conv layers may cause the loss of learning long-range contextual features. To make up for this loss, we design the first Transformer, ASVT, which adopts the idea of self-attention to aggregate information from both nearby and far-away voxels to better capture sparse voxel features. In ASVT, we define each individual voxel as an atom. During processing, each atom should be interacted with all other atoms according to the learned per-atom contributions. By doing so, different key atoms could be attended by other atoms so that both local relationship of nearby atoms and long-range contextual relationship of far way atoms will be learned, i.e, inter-atoms long-range contextual features are learned. Note that learning such kind of inter-atoms long-range contextual relationship  is very important for the model. For example, in a scene, assume there are two atoms that belong to different instances of the same category. If only SP-Conv is used, the "same-category" information may be ignored due to the small receptive field. While if AVST is added to learn such kind of information, the model can better encode what the scene describes. Hence the final global descriptor would be more powerful. The architecture of  ASVT is illustrated in Fig. \ref{ASVT}.

Let $X_{in} \in R^{L\times W \times H \times C}$ be the input sparse voxel features learned by SP-Convs (SP-voxel features for simplicity). We first learn the sparse voxel values (SP-values for simplicity) $X_v \in R^{L\times W \times H \times C}$, SP-queries $X_q \in R^{L\times W \times H \times C_r}$, and SP-keys $X_k \in R^{L\times W \times H \times C_r}$ through three different SP-Convs respectively:
\begin{equation}
\begin{split}
X_v=SPConv(X_{in})\\
X_q=SPConv(X_{in})\\
X_k=SPConv(X_{in})\\
\end{split}
\end{equation}
where we often set $C_r < C$ to reduce computational cost in later steps. That is to say, the dimension of  SP-queries and SP-keys are reduced from $C$ to $C_r$ for efficiency. After that, SP-voxel features of SP-values (SP-queries/keys) are rearranged to a tensor of $N \times C$ ($N \times C_r$), where $N$ is the number of non-empty voxels. The rearrange step is easy. Since coordinates and  features of non-empty voxels have been already stored as sparse tensors in SP-Conv's output, we only need to take out the feature tensor from its data structure for rearrange. Note that $N \ll L\times W \times H $ and $N \ll N_p $, where $N_p$ is point number of the raw point cloud, therefore, the SP-Conv and the following matrix multiplications based on the feature tensor are all very computational  efficient.

Then, we use $X_q$ and $X_k$ to calculate the SP-attention map $S$:
\begin{equation}
S=softmax(X_q \cdot X_k^T)
\end{equation}
where  $S \in R^{N \times N}$  encodes the contribution relationship of each atom with all the other atoms. In the following attending operation,  these relationships will contribute to aggregating  both short-range local information and long-range contextual information by interacting atoms. The attending operation can be summarized as:
\begin{equation}
X_s=SPConv(S \cdot X_v)
\end{equation}
where $X_s \in R^{N \times C}$ is called atom-attended SP-voxel features. In $X_s$,  features of each atom $x_i$ have adaptively accepted contributions from all the other atoms according to the implicit mode hidden in $S$. Thus meaningful contextual and semantic information can be represented in $X_s$ to describe the scene.

Finally, we rearrange  $X_s$ back to sparse voxel representations with a dimension of $L \times W \times H \times C $ and regard it as a residual term. The final ASVT feature is defined as the sum of $X_{in}$ and $X_s$:
\begin{equation}
X_{asvt}=X_{in}+X_s
\end{equation}

\begin{figure*}[t]
\centering  
\includegraphics[width=0.95\textwidth]{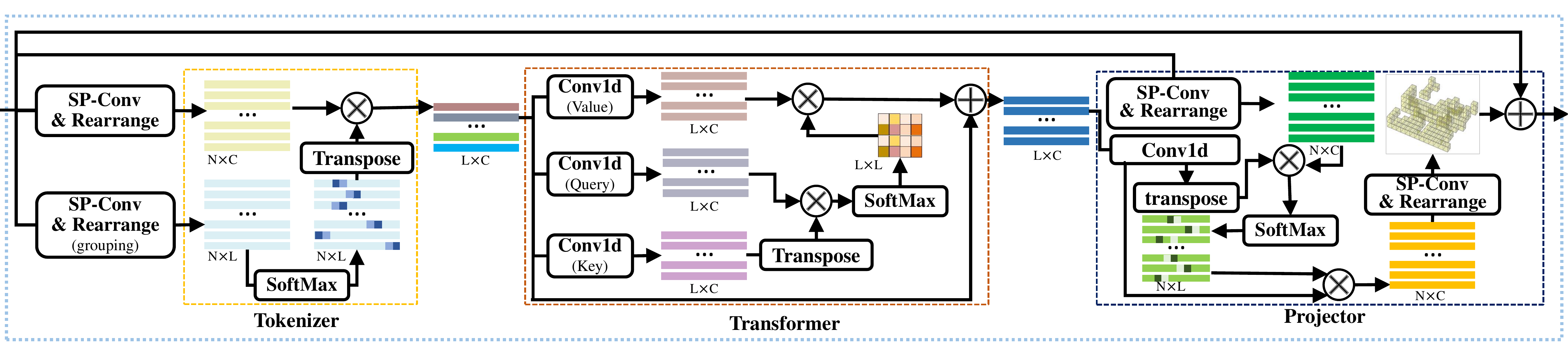}  
\caption{Network architecture of CSVT.} 
\label{csvt}
\end{figure*}

\subsection{Cluster-based Sparse Voxel Transformer}

Another observation we find is: in the sparse voxel representation, some atoms may share the same characteristics. For example, atoms of a wall always form a plane like structure, while atoms of a flower bed easily form a cylinder like structure. This means that atoms can actually cluster into different clusters according to their geometric or semantic characteristics, and the long-range contextual properties can also be extracted from the perspective of interacting between these clusters, i.e, learning inter-clusters long-range contextual features. Motivated by this intuition, we propose the second Transformer: CSVT. As shown in Fig. \ref{csvt}, CSVT consists of three component, a Tokenizer module, a Transformer module and a Projector module. They cooperatively learn how to implicitly group atoms into characteristics-similar clusters and interact clusters for enhancing learned features. Next, we will introduce them in detail. 

\textbf{The Tokenizer module} is used to transform the input SP-voxel features into tokens, where each token represents a cluster in the latent space.  We again define $X_{in} \in R^{L\times W \times H \times C}$  as the initial SP-voxel features. To achieve the goals of the tokenizer, we first use a SP-Conv operation followed by a rearrange operation to generate a grouping map $X_g \in R^{N \times L_t}$ :
\begin{equation}
X_{g}=softmax(RE(SPConv(X_{in})))
\end{equation}
where $RE$ is the rearrange operation. $L_t$ is the number of tokens we choose to generate and $N$ is the number of non-empty voxels. $X_{g}$  stores the probabilities of each voxel belonging to each token. Therefore, we can use $X_{g}$ to capture representations of tokens as grouping different tokens into different clusters in an implicit way:
\begin{equation}
T=X_{g}^T \cdot SPConv(X_{in})
\end{equation}
where $T \in R^{L_t \times C}$ denotes representations of $L_t$ tokens with each of them being described by $C$ features.

\textbf{The Transformer module} is then used to learn inter-clusters long-range contextual features through interacting these tokens. First, we generate values, keys, and queries using shared convolutional kernel Conv1d:
\begin{equation}
T_v=Conv1d(T), T_q=Conv1d(T), T_k=Conv1d(T)
\end{equation}
Then, tokens  are interacted with each other through the following attention operation:
\begin{equation}
T_s=T+Conv1d(softmax(T_q \cdot T_k^T) \cdot T_v)
\end{equation}
where $T_s \in R^{L_t \times C}$ is the attended tokens. Through the Transformer module, relationship between different clusters are learned to characterize the distribution characteristics of the scene with high quality. For example, the final descriptor may memorize that there is a rectangular building in the scene stands 5 meters away from a cylindrical building, or remember that there is a spherical building stands behind a slender tree.

\textbf{The Projector module} is then used to project token features back to the sparse voxel representation. Specifically, we use $T_s$ to calculate a re-projection map $M_p \in R^{N \times L_t}$:
\begin{equation}
T_p=Conv1d(T_s)
\end{equation}
\begin{equation}
M_p=\textrm{softmax}(RE(SPConv(X_{in})) \cdot T_p^T)
\end{equation}
where $T_p \in R^{L_t \times C}$. Then, the re-projection operation is defined as:
\begin{equation}
X_s=SPConv(M_p \cdot T_p)
\end{equation}
Again, we  rearrange  $X_s$ back to sparse voxel representations with a dimension of $L \times W \times H \times C $ and regard it as a residual term. The final CSVT feature is defined as:
\begin{equation}
X_{csvt}=X_{in}+X_s
\end{equation}

Note that, though both aim to learn long-range contextual features, roles and working mechanisms of ASVT and CSVT are different. The ASVT focus on learning relationship between similar and dissimilar individual atoms and learns inter-atoms long-range contextual features in a fine-grained level, while CSVT focus on learning relationship between different characteristics-similar clusters so that it learns inter-clusters long-range contextual features in a relative coarser level. They are complementary to each other.

\subsection{Network Architecture}

\begin{figure}[ht]
\centering  
\includegraphics[width=0.5\textwidth]{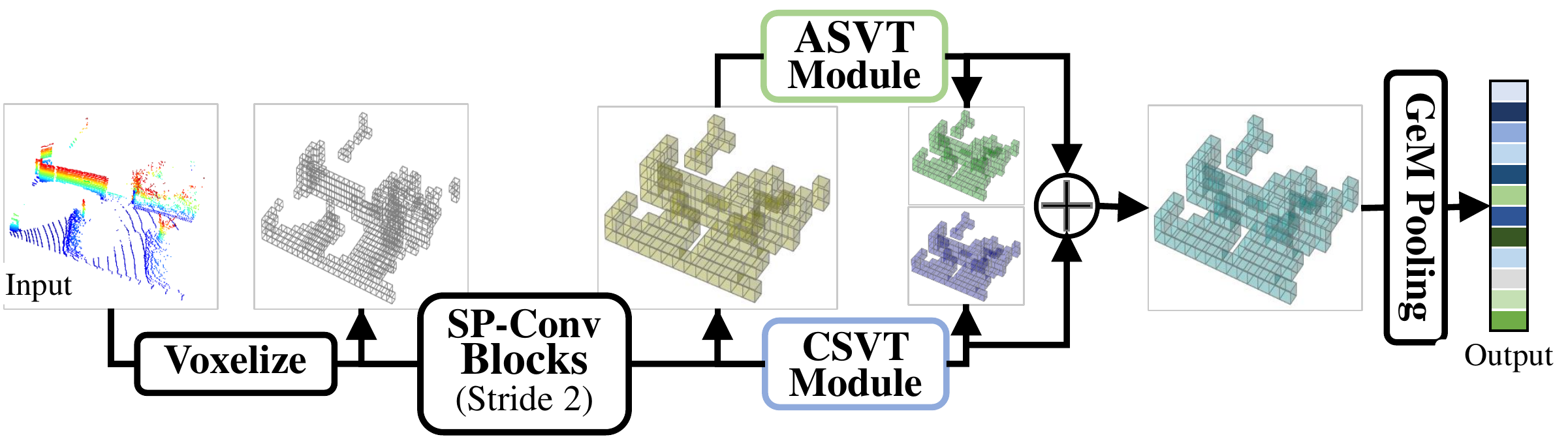}  
\caption{Pipeline of SVT-Net. The circle-add symbol means element-wise sum.} 
\label{pipeline}
\end{figure}

The overall architecture of SVT-Net is built upon the above introduced ASVT and CSVT as well as the light-weight SP-Conv. Specifically, as shown in  Fig. \ref{pipeline}. The initial sparse voxel representation is fed into the first SP-Conv layer with an output dimension of 32 to learn initial SP-features. Then two SP-Res-Blocks (each consists of two SP-Convs with a skip connection) are used to enhance learned features and increase the feature dimension to 64. Next, another SP-conv layer is used to increase the feature dimension to the final descriptor's dimension $d$. After that, the SP-features are fed into two branches for learning ASVT features and CSVT features using the two proposed Sparse Voxel Transformers(SVTs) respectively. Then, the learned ASVT features and CSVT features are fused by directly adding them together. Finally, the final global descriptor is calculated by using a GeM Pooling operation \cite{radenovic2018fine}:

\begin{equation}
f=[f_1, \cdot\cdot\cdot, f_k,\cdot\cdot\cdot, f_d],f_k=\frac{1}{|X_{final,k}|}\sum_{x \in X_{final,k}}{(x^{p_k})^{\frac{1}{p_k}}}
\end{equation}
where $f \in d$ is the final descriptor, $X_{final}$ is $X_{csvt}+X_{asvt}$, and $p_k$ is a learnable control parameter.

 Other details of the network architecture can be found in \textbf{Supp}. Thanks to the strong power of ASVT and CSVT, our proposed model SVT-Net can achieve superior performance compared to previous methods, even though our network architecture is simpler and smaller (from another words, it is shallower). Note that ASVT and CSVT can also be individually utilized  in different networks. Therefore, we propose two simplified versions of SVT-Net: ASVT-Net and CSVT-Net, by eliminating the ASVT module and CSVT module, respectively, to verify the effectiveness of the two modules. According to experimental results, both ASVT-Net and CSVT-Net also achieve state-of-the-art performances but further reduce the model size for a large margin.

\subsection{Loss Function}

In view of its superior performance in \cite{komorowski2021minkloc3d}, we adopt the following triplet loss to train our model:
\begin{equation}
L(f_i,f_i^p,f_i^n)=max\{d(f_i,f_i^p)-d(f_i,f_i^n)+m,0\}
\end{equation}
where $f_i$ is the descriptor of the query scan, $f_i^p$ and $f_i^n$ are descriptors of positive sample and negative sample respectively, and $m$ is a margin. $d(x,y)$ means the Euclidean distance between $x$ and $y$. To build informative triplets, we use batch-hard negative
mining following \cite{komorowski2021minkloc3d}.

After the network is trained, all point clouds are embedded into descriptors using the model. And we use the KNNs algorithm to find $m_s$  in the database, which is the closest one to the query scan $Q$.
\section{Experiments}
\subsection{Datasets and Metrics}

To fairly compare with other methods, we use the benchmark datasets proposed by \cite{uy2018pointnetvlad} to evaluate our method, which are now recognized as the most influential datasets for point cloud based place recognition. The benchmark contains four datasets: one outdoor dataset named Oxford generated from Oxford RobotCar \cite{maddern20171} and three in-house datasets: university sector (U.S.), residential area (R.A.) and business district (B.D.). The benchmark contains 21711, 400, 320, 200 submaps for training and 3030, 80, 75, 200 submaps for testing for Oxford., U.S., R.A. and B.D. respectively. Each point cloud contains 4096 points, which is the common setting of point cloud based place recognition. We use average recall at top 1\% and  average recall at top 1 as main metrics as previous methods for a fair comparison.


\begin{table*}[t]

\centering
\caption{Comparison with the state-of-the-art methods under the baseline setting.} 
\tabcolsep7pt
\begin{tabular}{l|cccc|cccc}
\hline
& \multicolumn{4}{c|}{{\bf Average recall at top-1\% (\%)}} &  \multicolumn{ 4}{c}{{\bf Average recall at top-1 (\%)}} \\
{\bf Method} & {\bf Oxford} & {\bf U.S.} & {\bf R.A.} & {\bf B.D.} & {\bf Oxford} & {\bf U.S.} & {\bf R.A.} & {\bf B.D.} \\
\hline
{\bf PointNetVLAD}  &       80.3 &       72.6 &       60.3 &       65.3 &          - &          - &          - &          - \\
{\bf PCAN}&       83.8 &       79.1 &       71.2 &       66.8 &          - &          - &          - &          - \\
{\bf DAGC} &       87.5 &       83.5 &       75.7 &       71.2 &          - &          - &          - &          - \\
{\bf SOE-Net} &       96.4 &       93.2 &       91.5 &       88.5 &          - &          - &          - &          - \\
{\bf SR-Net}  &       94.6 &       94.3 &       89.2 &       83.5 &       86.8 &       86.8 &       80.2 &       77.3 \\
{\bf LPD-Net}&       94.9 &         96.0 &       90.5 &       89.1 &       86.3 &         87.0 &       83.1 &       82.3 \\
{\bf Minkloc3D} &       97.9 &         95.0 &       91.2 &       88.5 &         93.0 &       86.7 &       80.4 &       81.5 \\
\hline
{\bf SVT-Net(Ours)} &       97.8 & {\bf 96.5} & {\bf 92.7} & {\bf 90.7} &       93.7 & {\bf 90.1} & {\bf 84.3} & {\bf 85.5} \\
{\bf ASVT-Net(Ours)} &   {\bf 98.0} &       96.1 &         92.0 &       88.4 & {\bf 93.9} &       87.9 &       83.3 &       82.3 \\
{\bf CSVT-Net(Ours)} &       97.7 &       95.5 &       92.3 &       89.5 &       93.1 &       88.3 &       82.7 &       83.3 \\
\hline
\end{tabular}  
\label{compare_baseline}
\end{table*}

\subsection{Implementation Details}

In all experiments, we voxelize 3D point coordinates with 0.01 quantization step. The voxelization and the following SP-Conv operation are performed by the MinkowskiEngine auto differentiation library \cite{choy20194d}. The dimension of the final descriptor is set to 256. The number of tokens $L_t$ is set to 8. Following previous work, we train two versions of models: baseline model and refined model. The baseline model is trained only using the training set of Oxford dataset, and the refined model is trained by adding the training set of U.S. and R.A. (Note that training set of B.D. is not added). Random jitter, random translation, random points removal and random erasing augmentation are adopted for data augmentation during training.  All experiments are performed on a Tesla V100 GPU with a memory of 32G. More details can be found in the \textbf{Supp}.


\subsection{Main Results}

In this section, we experimentally verify the effectiveness and efficiency of our method. Specifically, we first compare our models with PointNetVLAD \cite{uy2018pointnetvlad}, PCAN \cite{zhang2019pcan}, DAGC \cite{sun2020dagc}, SR-Net \cite{fan2020srnet}, LPD-Net \cite{liu2019lpd}, SOE-Net \cite{xia2021soe} and Minkloc3D \cite{komorowski2021minkloc3d} in terms of recognition accuracy. Then, we compare the inference time and model size between our models with them. Finally, we qualitatively analyze what the two SVTs have learned.

\begin{figure}[t]

\centering  
\includegraphics[width=0.45\textwidth]{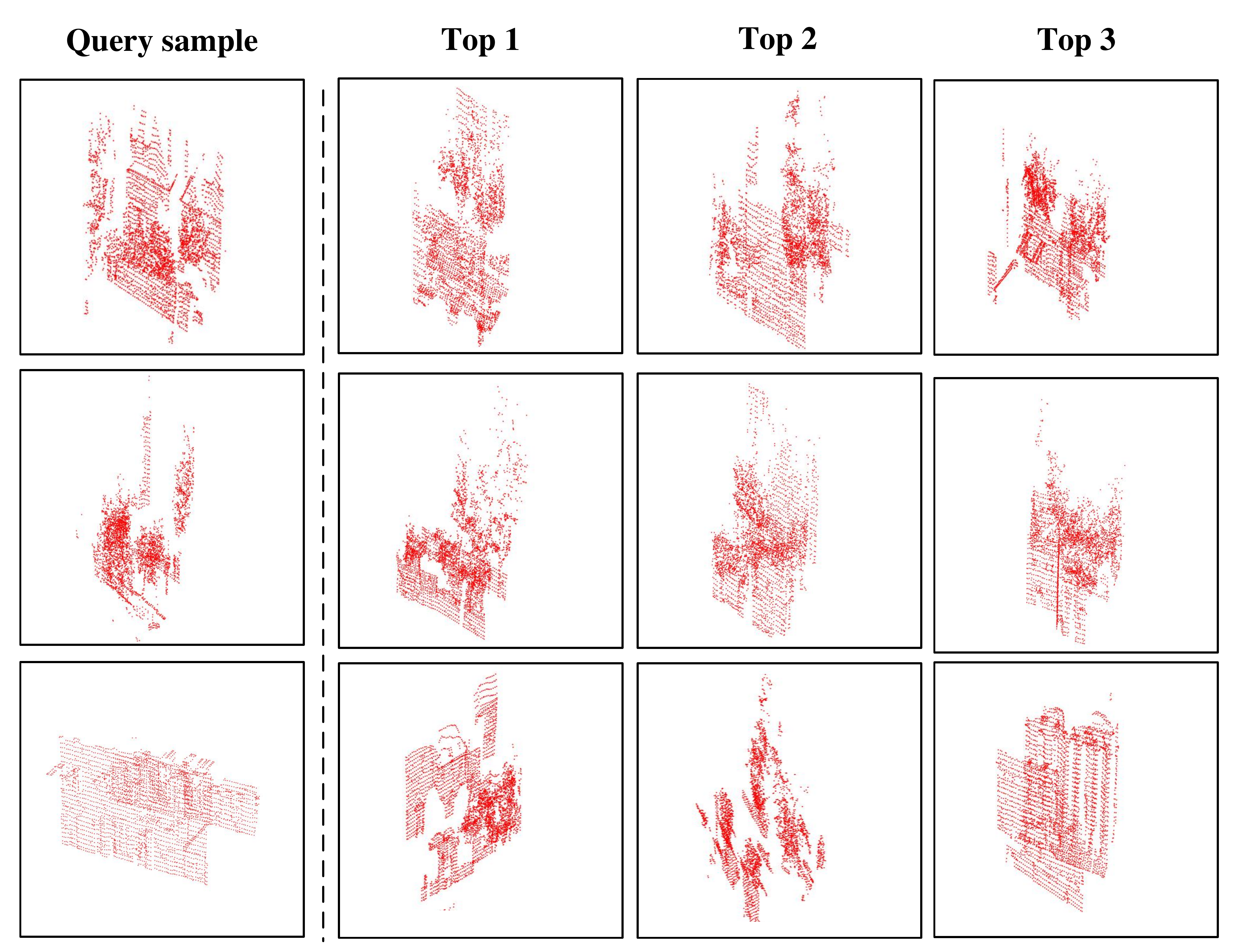} 

\caption{Visualization of top 3 matching results.}
\label{topk_visual}
\end{figure}

\textbf{Accuracy}:  In Table \ref{compare_baseline}, we show the results of all methods on the baseline setting. It can be found that SVT-Net significantly outperforms all state-of-the-art methods, especially for the average recall at top 1 metric on U.S., R.A., and B.D., where SVT-Net wins for 3.4\%, 3.9\%, 4\% compared to Minkloc3D respectively. Compared to SVT-Net, performances of ASVT-Net and CSVT-Net drop to some extent. However, their performances still largely outperform the previous best model Minkloc3D. We contribute the accuracy gain to the two novel SVTs we design. Note that Minkloc3D is also built upon SP-Conv and shares the same loss function as our model, while its performance is not as excellent as our models, which further confirms the superiority of our two proposed SVTs. Specifically, our SVT-Net build light-weight sparse voxel transformers
based on SPConv, while Minkloc3D simply stacks SPConv
layers, which is the main difference between Minkloc3D and our model, and therefore it is the two SVTs being the main force make our model perform better. What's more,  SOE-Net also use self-attention in its network architecture to learn long range context dependencies, but our model outperforms SOE-Net. This demonstrates that sparse voxel transformers are more effective than point-wise  transformers for large scale place recognition. We also note the self-attention module in SOE-Net is inefficient especially when the number of points is large due to computing attention weights for each of the $N_p$ raw points. In contrast, the novel ASVT and CSVT in our SVT-Net are built for processing sparse voxels, which are much more efficient because we only need to compute attention weights for each of $N$ ($N \ll N_p$) non-empty voxels. Recall curves of the baseline setting can be found in \textbf{Supp}.  We also visualize some top-k matching results in Fig. \ref{topk_visual} to provide readers with a comprehensive view to understand our place recognition results.

For a comprehensive comparison, we also show the results of all models at the refined setting in \textbf{Supp}. We find that at the refined setting, our models still significantly outperform all models except Minkloc3D. In fact, our models still perform better than Minklo3D in most cases, although only by a small margin. The difference between our three models becomes narrow. We attribute this to that all models have reached the performance upper bound.

\begin{table}
\centering
\caption{Efficiency comparison.} 
\begin{tabular}{l|c|c}
\hline
{\bf Method} & {\bf Time} & {\bf Parameters} \\
\hline
{\bf PointNetVLAD}  &          - &      19.8M \\
{\bf PCAN}  &          - &      20.4M \\
{\bf LPD-Net} &          - &      19.8M \\
{\bf Minkloc3D} &    12.16ms &       1.1M \\
\hline
{\bf SVT-Net(Ours)} &    12.97ms &       0.9M \\
{\bf ASVT-Net(Ours)} & {\bf 11.04ms} & {\bf 0.4M} \\
{\bf CSVT-Net(Ours)} &    11.75ms &       0.8M \\
\hline
\end{tabular}  

\label{speed_size} 
\end{table}

\begin{figure*}[t]

\centering  
\includegraphics[width=0.8\textwidth]{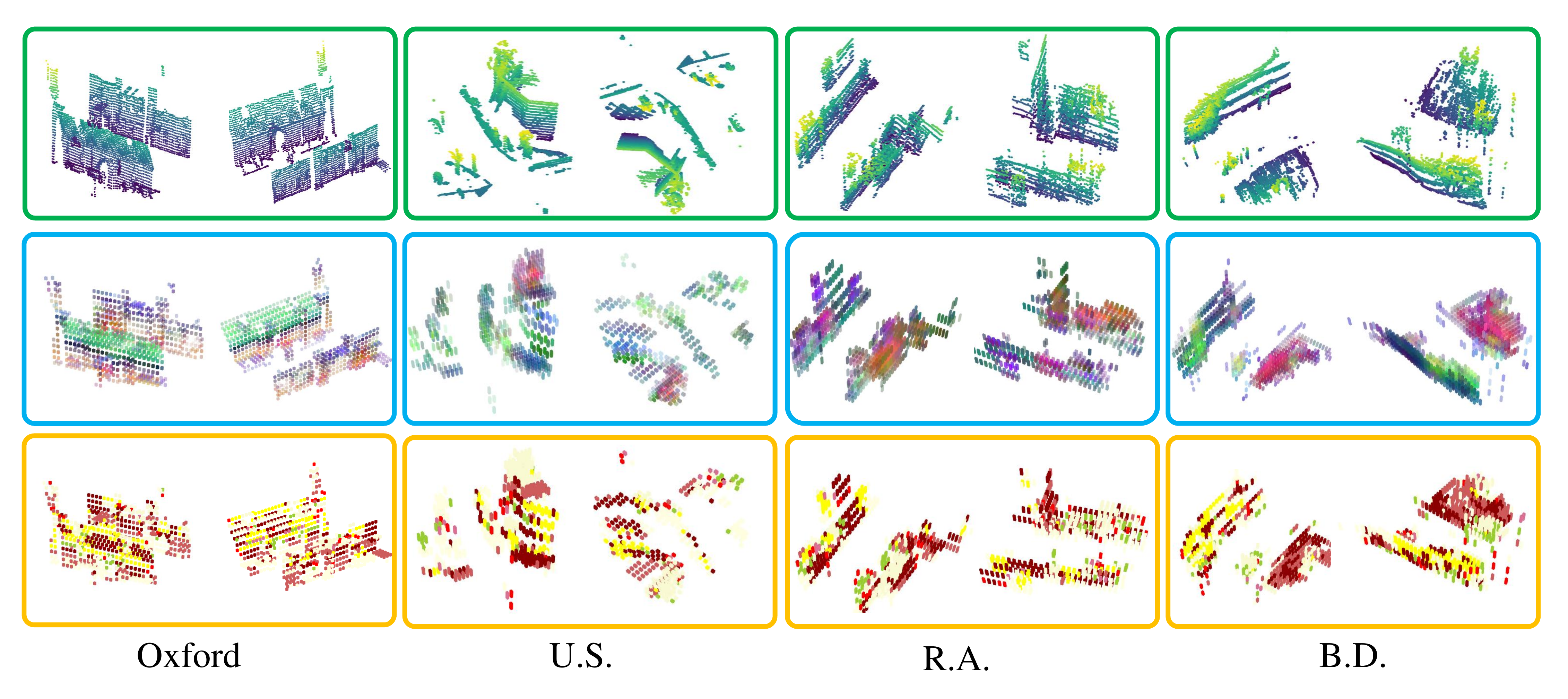} 

\caption{Visualization of what ASVT and CSVT have learned. First row: original point clouds. Second row: features learned by ASVT, "same category" atoms are attended similarly, e.g, in Oxford, two walls of the same height share the same color. Third row: features learned by CSVT,  atoms belong to the same geometric shape are clustered together and interacted with each other, e.g, in B.D., all the atoms in the same flowerbed (colored in Crimson) form a cube and are clustered together.}
\label{visualization}
\end{figure*}

\begin{table*}[t]
\centering
\caption{Results of ablation study for our SVT-Net.} 
\scalebox{0.95}{
\begin{tabular}{l|cccc|cccc}
\hline
& \multicolumn{4}{c|}{{\bf Average recall at top-1\%  (\%)}} &  \multicolumn{4}{c}{{\bf Average recall at top-1 (\%)}} \\
{\bf Method}  & {\bf Oxford} & {\bf U.S.} & {\bf R.A.} & {\bf B.D.} & {\bf Oxford} & {\bf U.S.} & {\bf R.A.} & {\bf B.D.} \\
\hline
{\bf A: $L_t$=4, $d$=256, $add$} &       97.9 &       96.4 &       92.5 &         89.0  &       93.7 &         89.0  &       83.9 &       82.5 \\
{\bf B: $L_t$=6, $d$=256, $add$} &   98.0  &       96.2 &       92.3 &       90.1 &       93.8 &       88.3 &       83.7 &       84.4 \\
{\bf C: $L_t$=10, $d$=256, $add$} &       97.9 &       96.2 &         92.0 &       89.4 &       93.8 &       87.2 &       83.3 &       83.5 \\
\hline
{\bf D: $L_t$=8, $d$=128, $add$} &       97.8 &       95.2 &         92.0 &         89.0  &       93.3 &       88.9 &       81.9 &       82.5 \\
{\bf E: $L_t$=8, $d$=384, $add$} & {\bf 98.2} &       94.8 &       92.5 &         89 & {\bf 94.4} &       86.9 & {\bf 84.9} &       83.7 \\
{\bf F: $L_t$=8, $d$=512, $add$} &         98.0  & {\bf 97.3} &       92.1 &       88.2 &       93.9 & {\bf 90.1} &         84.0  &       82.7 \\
\hline
{\bf G: $L_t$=8,$d$ =512, $cat$} &       97.5 &       93.4 &       85.8 &       84.7 &       92.7 &       81.9 &       73.9 &       77.1 \\
{\bf H: $L_t$=8, $d$=256, $cat\&spconv$} &       96.5 &       89.8 &       84.5 &       82.4 &       89.5 &       78.2 &       71.2 &         74.0 \\
{\bf SVT-Net: $L_t$=8, $d$=256, $add$} &       97.8 &       96.5 & {\bf 92.7} & {\bf 90.7} &       93.7 & {\bf 90.1} &       84.3 & {\bf 85.5} \\
\hline
\end{tabular}}

\label{ablation}
\end{table*}

\textbf{Model size and speed}: To verify the efficiency of our method, we compare our models with previous works in terms of model size and inference time in Table \ref{speed_size} and Fig. \ref{task_and_size} respectively.  For model size, it can be seen that SVT-Net and CSVT-Net save 18.2\% and 27.3\%  parameters respectively compared to the existing smallest model Minkloc3D. As for ASVT-Net, it even only has 36.4\% parameters of Minkloc3D, which is a significant reduction. And it is worth noting that all of our three models outperform Minkloc3D for a large margin in terms of accuracy at the baseline setting.  The ability of significantly improving accuracy under the condition of drastically reduced parameters  further fully demonstrates the superiority of our two SVTs. For speed,  compared to the current fastest model Minkloc3D, SVT-Net only add ignorable additional inference time. And  both ASVT-Net and CSVT-Net run faster than Minkloc3D. Approximately, voxelization and SP-Conv blocks cost about half of the running time, while ASVT and CSVT cost the another half. We find the speed increase is not as significant as the model size reduction, which is because that the inherent Transformer operation requires multiple matrix multiplications. Summing up the above results, we can conclude that our models are good enough in terms of both model size and running speed. Note, compared to Minkloc3D, our network architectures are much shallower, that's why our models are more light-weight than it. And since the main difference between our models with Minkloc3D is the two SVTs we design, we can contribute all performance gains into the learned long-range contextual features.

We believe that expect recognition accuracy, both storage efficiency and recognition accuracy are also significant factors to make solid and convincing comparisons. In this work, extensive results show that our model outperforms the SOTA in all the three aspects. Besides, we also find our three version  show different specialties towards the three different aspects, and so we  can accordingly make different utilization choice. Specifically, SVT-Net is larger than ASVT-Net and CSVT-Net, but performs better in most cases. Therefore, if there is enough resource, we recommend to use SVT-Net for the place recognition task.  If you only have limited computational resource and can’t fine-tune the model on new scenarios, we recommend to use CSVT-Net because its generalization ability towards new scenarios is better than ASVT-Net. Otherwise, ASVT-Net is a better choice because it is the fastest and the smallest one.

\textbf{What Transformers have learned}: One may be interested in what ASVT and CSVT have learned that could make our models so elegant. To explore this question, we show some visualization results in Fig. \ref{visualization}. The first row shows the original point clouds randomly selected from Oxford, U.S., R.A. and B.D respectively.

In the second row, we visualize the features of each non-empty voxel after ASVT using T-SNE \cite{van2008visualizing}. Different colors represent different distribution of these features in the feature space. It can be seen that by interacting each atom with all the others, the model indeed learns the relationship between atoms. Specifically, it is obvious that nearby atoms share the same color, which means they are attended similarly since they may belong to the same object parts. And it can be seen that far away atoms in the 3D space sharing the same implicit mode have similar colors, which means inter-atoms long-range features like relationship between far way  semantic similar atoms (e.g., the "same-category" information) has been discovered by the model.  A typical example that can prove the above analysis is: in Oxford, two walls of the same height share the same color, which means atoms of them are attend similarly. 

In the third row, we visualize which token that each non-empty voxel belongs to. Different color represents different tokens.  It can be seen that voxels belong to the same token always represent the same objects and share some common geometric characteristics. For example, in B.D., all the atoms in the same flowerbed (colored in Crimson) form a cube and are clustered together. This observation means that voxels indeed have been clustered together in the feature space according to their geometric characteristics. And obviously, the interaction between clusters or tokens could enhance model's understanding towards the scene. The inter-clusters long-range context properties like the relative positions between clusters would be encoded through such kind of interaction. In a word, the visualization results have confirmed our intuition of designing ASVT and CSVT and they have all contributed to the performance improvement.

\subsection{Ablation Study}

We have verified the effectiveness and efficiency of ASVT and CSVT in the \textbf{Main Results} section. Next, we experimentally study the effectiveness of other key designs. Specifically, we study the impact of the number of tokens $L_t$, dimension of descriptors $d$,  Transformer feature fusion strategy and training stability. We design experiments from A to H for this study. Table \ref{ablation} shows the results. "SVT-Net" in the last row of Table \ref{ablation}  refers to the model we finally choose.

\textbf{Impact of number of tokens}: The number of tokens ($L_t$) decides how many clusters we divide the scene into. We change the value of $L_t$ and compare the results in  Table \ref{ablation}.  Comparing the experiment A, B, C and SVT-Net, we find that setting $L_t$ as 8 is the best choice. When $L_t$ is too small, interaction between different geometric characteristic (hidden in different clusters) would  be limited. When $L_t$ is too large, it is easy to cause over-fitting.

\textbf{Impact of descriptor's dimension}: To a certain extent, the dimension $d$ determines the  descriptor's capability  of  describing a scene. From experiment D, E, F and SVT-Net in Table \ref{ablation}, we find that overall larger dimension leads to better performance. However, when it is larger than 256, the accuracy increase is minimal while the model size is significantly increased to 1.8M and 3.0M for $d=384$ and $d=512$ respectively. Therefore, for a better trade-off between accuracy and model size, we choose $d=256$ in our implementation.

\textbf{Impact of fusion strategy}: In SVT-Net, we need to fuse features learned by ASVT and CSVT before aggregating voxel features into a global descriptor. In experiment G, we investigate the effectiveness of another fusion method, concatenation. In this way, the output dimension is $512$. However, the performance of concatenating the two features is not as good as simply adding them (the dimension is 256). Then, we suspect if it is the higher dimension that causes the performance drop. Therefore, in experiment $H$, we add an additional SP-Conv layer after  concatenation to make the dimension be 256. Unfortunately, the model's performance  becomes even worse than before. Therefore, finally, we believe that direct adding together is the best way to fuse the features of the two SVTs.

\textbf{Training stability}: We notice that for each training time, there are some small differences on the evaluation results. To avoid conclusion bias, we train each model for multiple times and  show the boxplot of each model in \textbf{Supp}, which reflects the training stability of each model. Considering the trade off between accuracy, model size, and training stability, we claim that SVT-Net is the best performed model.

\section{Conclusions}
In this paper, we  proposed a super light-weight network  for large scale place recognition named SVT-Net. In SVT-Net, two Sparse Voxel Transformers: Atom-based Sparse Voxel Transformer (ASVT) and Cluster-based Sparse Voxel Transformer (CSVT) are proposed to learn long-range contextual properties. Extensive experiments have demonstrated that SVT-Net as well as its two simplified versions ASVT-Net and CSVT-Net can all achieve state-of-the-art performance with an extremely light-weight network architecture. 

\textbf{Acknowledgements}  This work was supported in part by National Key Research and Development Program of China under Grant No. 2020YFB2104101 and National Natural Science Foundation of China (NSFC) under Grant Nos. 62172421, 71771131 and U1711262.

{\small
\bibliographystyle{aaai22}
\bibliography{aaai22}
}
\newpage
\section{Supplementary Material}
\textbf{Implementation details}: In all experiments, we voxelize 3D point coordinates with 0.01 quantization step. The voxelization and the following SP-Conv operation are performed by the MinkowskiEngine auto differentiation library. The dimension of the final descriptor is set to 256. The number of tokens $L_t$ is set to 8. Following previous work, we train two versions of models: baseline model and refined model. The baseline model is trained only using the training set of Oxford dataset, and the refined model is trained by adding the training set of U.S. and R.A. (Note that training set of B.D. is not added). Random jitter, random translation, random points removal and random erasing augmentation are adopted for data augmentation during training.  All experiments are performed on a Tesla V100 GPU with a memory of 32G.

In ASVT, dimension of $X_q$ and $X_k$ is reduced by a factor of 8 from the input, i.e. from $256 \to 32$. The margin $m$ in the loss function is set to 0.2. To prevent embedding collapse in early epochs of training, we use a dynamic batch sizing strategy. During training, we count the number of active triplets, when it falls below 70\% of the current batch size, the batch is increased by 40\% until the maximum size of 256 elements is reached. In the baseline setting, the initial batch size is 32 and the initial learning rate is $10^{-3}$. The model is trained for 40 epochs and the learning rate is decayed by 10 at the end of the 30th epoch.  The refined model is trained with an  initial batch size of 16 and an initial learning rate of $10^{-3}$. The model is trained for 80 epochs and the learning rate is decayed by 10 at the end of the 60th epoch. The model is implemented by Pytorch and optimized by Adam optimizer. In training, point clouds are regarded as correct matches if they are at maximum 10m apart and wrong matches if they are at least 50m apart. In testing, the retrieved point cloud is regarded as a correct match if the distance is within 25m between the retrieved point cloud and the query scan.

\textbf{Details about model architecture}:  We show the detailed information of our  model architecture in Table \ref{tab:svtnet_parameters}. Our SVT-Net takes the sparse voxel representation of a point cloud as input. The input is first processed by a SP-Conv with a kernel size of $5 \times 5$ and the stride is 1. Then, two ResNet-like operations are used to learn  local features, each operation consists of a  $3 \times 3$  SP-Conv with stride 2 and a  $3 \times 3$  SP-Resblock  with stride 1. Then a $1 \times 1$ SP-Conv is used to lift the feature dimension to 256. Next, ASVT and CSVT are used separately for learning different long-range contextual features. Finally, a GeM-Pool is utilized for learning the final descriptor. Parameter numbers of each block are also shown in Table \ref{tab:svtnet_parameters}. The total parameter number of SVT-Net is about 0.936M. 

  \textbf{Results of the refined model}: For a comprehensive comparison, we  show the results of all models at the refined setting in Table \ref{compare_refine}. We find that at the refined setting, our models still significantly outperform all models except Minkloc3D.  In fact, our models still performs better than Minklo3d  in most cases, although only by a small margin. The difference between our three models becomes narrow. We attribute this to that all models have reached the performance upper bound. This observation also motivates us to build a more large scale dataset to more fairly evaluate current algorithms and promote future researches. And we leave it as our future work.

\begin{figure*}[ht]
\centering
\includegraphics[width=\textwidth]{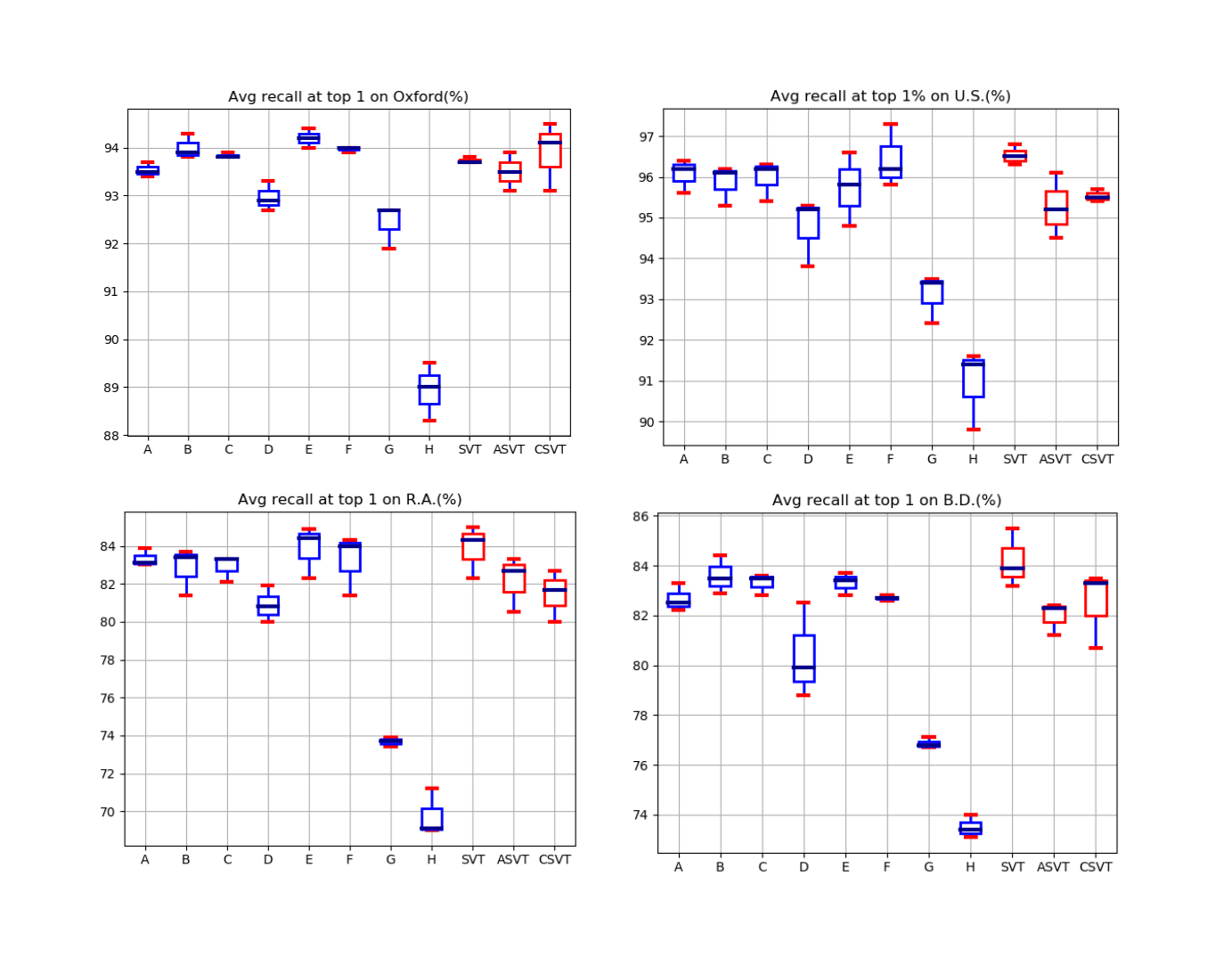}
\caption{Visualization of training stability. }
\label{boxplot}
\vspace{-1in}
\end{figure*}

 \textbf{Training stability}: We notice that for each training, there are some small differences on the evaluation results. To avoid conclusion bias, we train each model for multiple times and  show the boxplot of each model in Figure \ref{boxplot}, which reflects the training stability of each model.  It can be found in the figure that among all settings, the final version of SVT-Net achieves the best trade-off between accuracy, model size, and training stability. Therefore, we claim that SVT-Net is the best performed model.

\begin{figure*}[ht]
\centering  
\includegraphics[width=\textwidth]{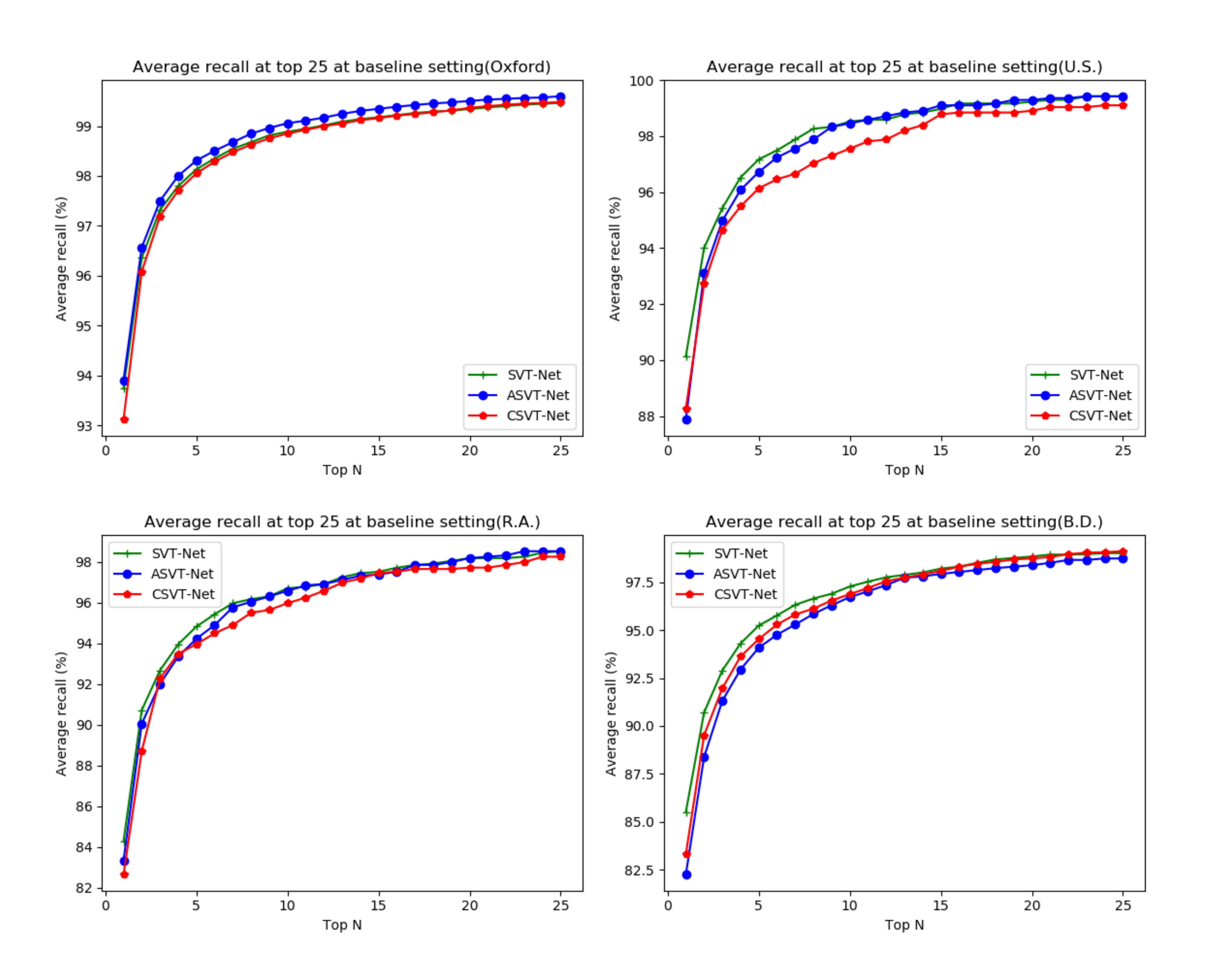}  
\caption{Recall curves of the baseline setting.} 
\label{recallcurve}
\end{figure*}

\textbf{Average recall at top N}: We show the recall curve of our three models at the baseline setting in Figure \ref{recallcurve}, which reflects the performance of the models about average recall at top N. It can be found that our three models have already performed very well at top 1. And if we relax the constraints, performance of models largely increase. It demonstrates that our models have the potential of meeting the practical requirements of SLAM systems in terms of accuracy.

\textbf{Limitation and future work}: There are still some limitations that have to be improved. First, the current work doesn't consider how to handle complex situation such as when the point cloud is sparse. Second,  performances of ASVT and CSVT on other tasks are unclear. In the future, we will investigate how to solve the two issues.


\begin{table*}
	\caption{The SVT-Net Parameters}
	\small
	\centering
	\begin{tabular}{ccccccc}
		\toprule
		{\multirow{2}*{\bf Blocks \& Layers} } & \multicolumn{2}{c}{\bf kernel} & \multicolumn{2}{c}{\bf channels}  & \\ 
		\cmidrule(lr){2-3} 
		\cmidrule(lr){4-5}
		~ & size & stride & in & out & \# params & input \\ 
		\midrule
		 conv0  & 5 & 1 & 3   & 32  & $\approx$ 4.0K & sparse voxel \\ \midrule 
		 convs[0]  & 2 & 2 & 32  & 32   & $\approx$ 8.1K  & conv0      \\ \midrule 
		 resblocks[0] & 3 & 1 & 32 & 32 & $\approx$ 55.4K  & convs[0]  \\ \midrule
		 convs[1]  & 3 & 2 & 32  & 32  & $\approx$ 8.1K  & resblock[0]      \\ \midrule 
		 resblocks[1] & 3 & 1 & 32 & 64 & $\approx$ 168.3K  & convs[1]  \\ \midrule
		 conv1x1 & 1& 1& 64 & 256 &  $\approx$ 16.3k & resblocks[1] \\ \midrule
		 asvtblocks  & 1 & 1 & 256  & 256 & $\approx$ 147.9K  & conv1x1      \\ \midrule 
		 csvtblocks  & 1 & 1 & 256 & 256  & $\approx$ 526.8K & conv1x1  \\\midrule 
		  GeM Pool  & -  & -  & - & -  & $ $ 1 & csvtblocks+asvtblocks\\  
		\bottomrule
		Total Parameters & $\approx$ 0.936M
	\end{tabular}
	\label{tab:svtnet_parameters}

\end{table*}

\begin{table*}[ht]
\centering
\caption{Comparison with the state-of-the-art methods under the refined setting.} 
\begin{tabular}{l|cccc|cccc}
\hline
& \multicolumn{4}{c|}{{\bf Average recall at top-1\% (\%)}} &  \multicolumn{4}{c}{{\bf Average recall at top-1 (\%)}} \\
{\bf Method}  & {\bf Oxford} & {\bf U.S.} & {\bf R.A.} & {\bf B.D.} & {\bf Oxford} & {\bf U.S.} & {\bf R.A.} & {\bf B.D.} \\
\hline
{\bf PointNetVLAD}  &       80.1 &       90.1 &       93.1 &       86.5 &       63.3 &       86.1 &       82.7 &       80.1 \\
{\bf PCAN}&       86.4 &       94.1 &       92.3 &         87.0 &       70.7 &       83.7 &       82.3 &       80.3 \\
{\bf DAGC}  &       87.8 &       94.3 &       93.4 &       88.5 &       71.4 &       86.3 &       82.8 &       81.3 \\
{\bf SOE-Net}  &       96.4 &       97.7 &       95.9 &       92.6 &       89.3 &       91.8 &       90.2 &         89.0 \\
{\bf SR-Net}  &       95.3 &       98.5 &       93.6 &       90.8 &       88.5 &       93.5 &       86.8 &       85.9 \\
{\bf LPD-Net} &       98.2 &       98.2 &       94.4 &       91.6 &         93.0 &       90.5 &       {\bf 97.4} &       85.9 \\
{\bf Minkloc3D} &       98.5 &       99.7 &       99.3 &       96.7 & {\bf 94.8} &       97.2 &  96.7 &         94.0 \\
\hline
{\bf SVT-Net(Ours)} &       98.4 & {\bf 99.9} & {\bf 99.5} &       97.2 &       94.7 &         97.0 &       95.2 &       94.4 \\
{\bf ASVT-Net(Ours)} &       98.3 &       99.6 &       98.9 &         97.0 &       94.6 & {\bf 97.5} &         95.0  & {\bf 94.5} \\
{\bf CSVT-Net(Ours)} & {\bf 98.6} &       99.8 &       98.7 & {\bf 97.3} & {\bf 94.8} &       96.6 &       96.2 &       94.3 \\
\hline
\end{tabular}  
\label{compare_refine}
\end{table*}

\end{document}